\documentclass{edm_article}
\usepackage[ 
  colorlinks=false,
  pdfborder={0 0 0},
  linkcolor=red
]{hyperref}
\usepackage[capitalize]{cleveref}
\usepackage{array} % add this package to the preamble
\usepackage{algorithm}
\usepackage{algorithmicx}
\usepackage{algpseudocode}
\begin{document}

\title{Understanding Revision Behavior in Adaptive Writing Support Systems for Education}

% Submissions for EDM are double-blind: please do not include any author names or affiliations in the submission. 
% Anonymous authors:
\numberofauthors{4}

%An example of how to include
% multiple authors is below for after the paper has been accepted.

% You need the command \numberofauthors to handle the 'placement
% and alignment' of the authors beneath the title.
%
% For aesthetic reasons, we recommend 'three authors at a time'
% i.e. three 'name/affiliation blocks' be placed beneath the title.
%
% NOTE: You are NOT restricted in how many 'rows' of
% "name/affiliations" may appear. We just ask that you restrict
% the number of 'columns' to three.
%
% Because of the available 'opening page real-estate'
% we ask you to refrain from putting more than six authors
% (two rows with three columns) beneath the article title.
% More than six makes the first-page appear very cluttered indeed.
%
% Use the \alignauthor commands to handle the names
% and affiliations for an 'aesthetic maximum' of six authors.
% Add names, affiliations, addresses for
% the seventh etc. author(s) as the argument for the
% \additionalauthors command.
% These 'additional authors' will be output/set for you
% without further effort on your part as the last section in
% the body of your article BEFORE References or any Appendices.

% \numberofauthors{8} %  in this sample file, there are a *total*
% % of EIGHT authors. SIX appear on the 'first-page' (for formatting
% % reasons) and the remaining two appear in the \additionalauthors section.
% %
\author{
% % You can go ahead and credit any number of authors here,
% % e.g. one 'row of three' or two rows (consisting of one row of three
% % and a second row of one, two or three).
% %
% % The command \alignauthor (no curly braces needed) should
% % precede each author name, affiliation/snail-mail address and
% % e-mail address. Additionally, tag each line of
% % affiliation/address with \affaddr, and tag the
% % e-mail address with \email.
% %
% % 1st. author
\alignauthor
 Luca Mouchel\\
        \affaddr{EPFL}\\
        \email{luca.mouchel@epfl.ch}
% % 2nd. author
 \alignauthor
 Thiemo Wambsganss\\
        \affaddr{EPFL}\\
        \email{thiemo.wambsganss@epfl.ch}\\
% % 3rd. author
 \alignauthor 
 Paola Mejia-Domenzain\\
        \affaddr{EPFL}\\
        \email{paola.mejia@epfl.ch}\\
\and
% use '\and' if you need 'another row' of author names
% % 4th. author
 \alignauthor Tanja Käser\\
        \affaddr{EPFL}\\
        \email{tanja.kaeser@epfl.ch}
}
% % There's nothing stopping you putting the seventh, eighth, etc.
% % author on the opening page (as the 'third row') but we ask,
% % for aesthetic reasons that you place these 'additional authors'
% % in the \additional authors block, viz.

% Just remember to make sure that the TOTAL number of authors
% is the number that will appear on the first page PLUS the
% number that will appear in the \additionalauthors section.

\maketitle

\begin{abstract}
Revision behavior in adaptive writing support systems is an important and relatively new area of research that can improve the design and effectiveness of these tools, and promote students' self-regulated learning (SRL). Understanding how these tools are used is key to improving them to better support learners in their writing and learning processes. In this paper, we present a novel pipeline with insights into the revision behavior of students at scale. We leverage a data set of two groups using an adaptive writing support tool in an educational setting. With our novel pipeline, we show that the tool was effective in promoting revision among the learners. Depending on the writing feedback, we were able to analyze different strategies of learners when revising their texts, we found that users of the exemplary case improved over time and that females tend to be more efficient. Our research contributes a pipeline for measuring SRL behaviors at scale in writing tasks (i.e., engagement or revision behavior) and informs the design of future adaptive writing support systems for education, with the goal of enhancing their effectiveness in supporting student writing. The source code is available at \href{https://github.com/lucamouchel/Understanding-Revision-Behavior}{\texttt{https://github.com/lucamouchel/\\Understanding-Revision-Behavior}}.
\end{abstract}

 % Replace with your own 3-5 keywords

\section{Introduction}
%Motivation 4-5 sentences about "Why are writing support tools important for learning"
Intelligent writing support tools (e.g., Grammarly, WordTune or Quilbot) offer new ways for learners to receive feedback and thus revise their texts \cite{nova2018utilizing}. These and other writing support systems bear the potential to provide learners with needed adaptive feedback on their writing exercises when educators are not present, (e.g., on grammatical mistakes \cite{UCAM-CL-TR-904}, argumentation \cite{WAMBSGANSS2022104644}, empathy \cite{wambsganss2021supporting}, or general persuasive writing \cite{wambsganss2022alen}). They can help students in their self-regulated learning (SRL) process \cite{Bandura,zimmerman}, to organize their thoughts and ideas, reflect on their learnings, or simply receive feedback on frequently occurring grammar or argumentation mistakes.
From an educational perspective, it is important to understand how these tools are used by learners in educational settings and how they improve the effectiveness of educational scenarios \cite{CHEN201859, flippedclassroms}. Present research is largely focused on designing and building writing support systems \cite{DBLP:coenen,DBLP:padmakumar}. However, there are not many insights into the effects of the usage of these tools and their impact on students' SRL processes \cite{bjork2013self, vandermeulen2020reporting}, which is why we contribute a novel pipeline analyzing and visualizing revision behavior to better understand how we can design, develop and improve existing systems to better support students.
%Solution: Using educational data mining to trakc and anaylise key stroke behavior. cite 4-5 related work on what has been done and then say whats novel about our paper for the solution of this problem
Techniques from the field of data mining are a solution to understanding revision behavior and explaining SRL. One such technique is Keystroke Logging (KL). KL allows us to use educational data mining to analyze user behavior in writing tasks \cite{leijten_keystroke_2013, zhu_analysis_2019, sinharay_prediction_2019}\footnote{Tools such as InputLog \cite{leijten_keystroke_2013} or ETS \cite{zhu_analysis_2019} are examples of KL programs.}. In this study, we model, inspect, and analyze quantitative data in learners' writing interactions through KL by developing a novel pipeline. We use a keystroke log from an experiment, where users were divided into two groups. The first one was given adaptive feedback and the second one was not. A detailed description of our dataset and the experiment demographics and procedure are available in \cref{sec:experiment}. To the best of our knowledge, no publicly available pipeline exists that focuses on processing the keystroke behavior of learners and helps analyze SRL characteristics such as engagement, revision, or visualize the learning path. We intend to first identify and visualize the differences between these two groups in their revision process and compare different user profiles and measure their engagement over time. We use an exemplary data set to build this pipeline and apply data mining in order to gain insights into the underlying process of this writing activity. 

\section{Background} 

\subsubsection*{Research on Automatic Data Mining for Writing Behaviour}

Research in writing process analysis can be traced to the 1970s \cite{emig1971composing, stallard1974analysis}. However, only more recently have studies been focusing theoretically on behavioral and cognitive processes of writing \cite{hayes2012modeling, mccutchen1996capacity}. In fact, Flower and Hayes \cite{10.2307/356600} laid the groundwork for research on the psychology of writing. They propose that the act of writing is propelled by goals, which are created by the writer and grow in number as the writing progresses. Today, writing support tools need to support this cognitive process as it emphasizes writers' intentions, rather than their actions \cite{gero-etal-2022-design}. It is important to understand what these tools help with, and how we may design new ones \cite{gero-etal-2022-design}. While prior works on text revision \cite{DBLP:coenen, 2022arXiv220106796L, DBLP:padmakumar, wambsganss__2020} have proposed machine collaborative writing interfaces, they focus on collecting human-machine interaction data to better train neural models, rather than understanding the underlying processes of text revision.
Several studies in the past have used KL as a technique to study revision \cite{leijten_keystroke_2013, sinharay_prediction_2019, zhu_analysis_2019} in different settings and some of the aims were to understand and evaluate keystroke log features in a writing task context. However, until now, KL has been scarcely used in the classroom \cite{vandermeulen2020reporting}. One issue with keystroke loggers is their invasive nature. KL raises several ethical issues, most notably privacy violation \cite{vandermeulen2020reporting}, but in this study, participants gave their consent for the collection of their data, all the while preserving their privacy.
Previous research has suggested that writing time and number of keystrokes, which are indicative of general writing fluency and effort, are related to writing quality \cite{allen2016enter, zhang2016classification}. Another feature of interest is pause times, \cite{zhang2017investigation} found that under a certain timed-writing test condition, shorter pauses are preferred as that indicates an adequate understanding of the task requirements, more familiarity with the writing topic, and better task planning \cite{sinharay_prediction_2019}.

\subsubsection*{Self-Regulated Learning}
To analyze revision behavior, we rely on the lens of self-regulated learning (SRL). 
SRL refers to the pro-active process that learners engage in to optimize their learning outcome \cite{zimmerman}. According to Zimmerman's model of SRL \cite{zimmerman}, there are three major phases: forethought, performance and self-reflection.
The forethought phase includes task analysis, such as goal setting and strategic planning and self-motivational beliefs. The performance phase includes self-control processes, such as task and attention-focusing strategies. The self-reflection phase includes processes involving self-judgment and self-reaction \cite{WONG2021106596}.
SRL is essential in the context of studying revision behavior in writing support systems as it allows writers to take an active role in identifying and addressing their own writing weaknesses, rather than simply relying on the writing support system to automatically detect and correct errors. This can lead to a deeper understanding of the writing process.

\section{Method}
%Contributions 4-5 sentences

\begin{figure}[!h]
\Description{Overview of our pipeline and methodology, following the KDD process}
  \centering
  \includegraphics[width=1\linewidth]{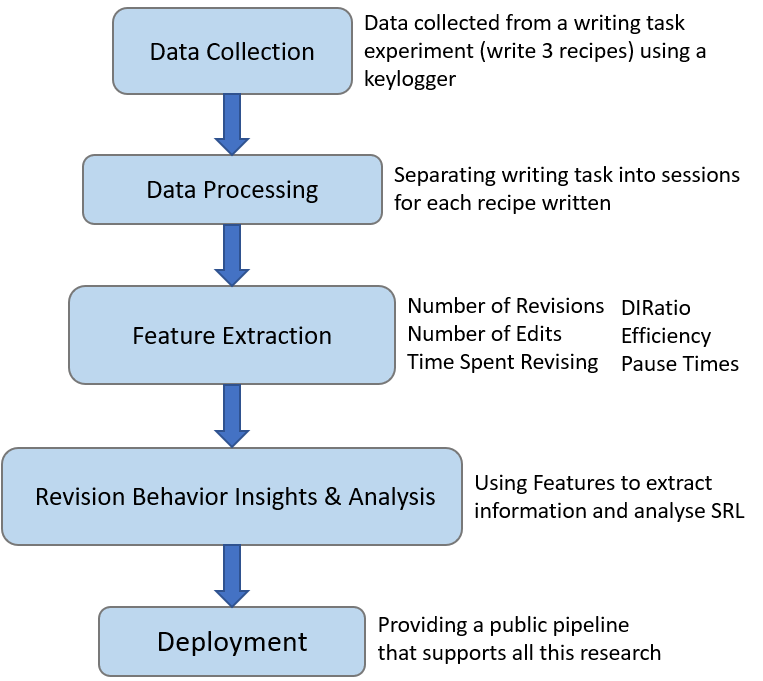}
  \caption{Overview of our pipeline and methodology, following the KDD process}
  \label{fig:method}
\end{figure}

To investigate revision behavior in the writing process, we propose a pipeline for the automatic analysis of the SRL behavior of users during a writing task. Our work follows the Knowledge Discovery in Databases process by following the methodology in \cref{fig:method}.

\subsubsection*{Demographics, Procedure \& Dataset Description}
\label{sec:experiment}
With approval of the ethical board of our university, we collected data from a writing experiment which consisted of 73 users divided into two groups, as illustrated in Table \ref{Table:experiment}.

\begin{table}[!h]
\centering
\renewcommand{\arraystretch}{1.3} % Increase the array stretch to add space between rows
\caption{Demographics of the participants per group from the exemplary data set}
\begin{tabular}{l|>{\centering\arraybackslash}p{2.2cm} >{\centering\arraybackslash}p{2.7cm}}
& With Adaptive Feedback (\texttt{G1}) & Without Adaptive Feedback (\texttt{G2})\\
\hline
No. Participants &  34 & 39 \\
Age Mean &  26.8 &  26.3\\
Age Std &  3.3 &  2.8\\
\hline
\% Female & 43 & 51\\
\% Male & 51 & 46\\
\% Other &  6 &  3\\
\end{tabular}
\label{Table:experiment}
\end{table}

 The two groups of users were tasked with writing three cooking recipes. Both groups were given a sample recipe as reference. The first group (\texttt{G1}) received adaptive feedback from the platform when they submitted their texts. The second group (\texttt{G2}) did not receive any feedback. Once they submitted their recipes to the system, users in \texttt{G1} had the option to reset and start a new recipe or revise their texts based on the feedback. The same protocol was followed for \texttt{G2}, but they did not receive feedback. Here are several examples of the feedback the platform provided users: \textit{'List each ingredient separately.'}, \textit{'Enumerate the steps.'}, \textit{'How can your recipe be more specific?'}, \textit{'Use stir, mix, or beat instead of “add” to be more specific.'} or \textit{'Indicate whether the meat, poultry, or seafood is boned, skinned, or otherwise prepared.'}

With regards to the dataset, the entries of the log data we collected consisted of user ids, event dates, the keystroke logs as a JSON file and the final version of the text submitted at that particular date. An example of an entry is as follows: \texttt{2023-01-01, 12:00:00, user1, [\{'time': 1, 'character': 'a'\}, ... \}], "a) Cook ..."}.
\subsubsection*{Qualitative Perception}
\label{subsec:quality}
Following the experiment, users were tasked with answering follow-up questions and we identified eight different topics regarding the reported revisions, including, \textit{adding missing ingredients}, \textit{improving the clarity}, \textit{not making any changes} and others. To do this, we used \texttt{BERTTopic} \cite{grootendorst2022bertopic}, a topic modeling technique that clusters sentence embeddings generated by \texttt{Sentence-BERT} \cite{DBLP:conf/emnlp/ReimersG19}, to perform qualitative analysis of participants' open responses about recipe revisions: (\textit{What did you edit (add, remove or change) from the original text (the recipe you wrote)?}). We split the sentences into clusters based on their relevance, assigned names to each cluster, and computed the probability of each sentence belonging to a cluster. We grouped the sentences by participant to obtain the set of topics associated with their entire text answer. For example, if a participant's answer consisted of sentences with assigned topics $A$, $B$, and $C$, the set of topics associated with their answer would be $Z = \{A, B, C\}$.

\subsubsection*{Data Processing}
\label{data processing} 
Given that the logs consist of the users' first attempts at writing one of the three recipes and their respective revision phases, it is important to separate them in order to focus only on the revision steps.
We define sessions for a user as all the data collected from them for \textbf{one} recipe. To separate sessions, we use cosine distance to detect where the session ends and where the next one starts. One advantage of using cosine distance for text comparison is that it is relatively insensitive to the length of the strings. In contrast, other measures of distance such as Euclidean distance are sensitive to the length of the vectors and can be affected by the presence of common words that do not contribute significantly to the meaning of the strings. To map sentences to 50-dimensional vectors, we use a GloVe model \cite{pennington2014glove}, which is already trained on Wikipedia. First, we map each word to their embeddings and then compute the sum of the vectors component-wise. Formally, each text submitted $t$ has a set of words $\mathcal{W} = \{w_i\; |\; 1 \leq i \leq N_t\}$, where $N_t$ is the number of words for text $t$. Then, we map each $w_i$ to their embeddings $\tilde w_i$ which are 50-dimensional vectors. Now let $\tilde{t}$ be the embedding of the text $t$, then $\tilde {t}= \sum_{i=1}^{N_t} \tilde w_i$. This allows us to capture each word of the text and this way, we can collect the set of all text embeddings in the dataset $\mathcal{T} = \{\tilde t_1, \tilde t_2, ...\} = \left\lbrace\sum_{i=1}^{N_1} \tilde w_i, \sum_{i=1}^{N_2} \tilde w_i, ...\right\rbrace$. We use $\mathcal{T}$ to run the recursive algorithm described in \cref{sec:writingsessions} on the recipes submitted and compute the cosine distance between the text embeddings of $t_k$, $t_{k+1}$, $...$, starting at $k=0$, until we find $n>k$ 
$$1-\frac{\langle \tilde t_k, \; \tilde t_n \rangle}{\lVert\tilde t_k\rVert\cdot \lVert\tilde t_n
\rVert} < 0.995$$
When we do, we define $n$ as the index of a new recipe in our data. Then, we repeat the process by starting at $t_n$ and comparing $\tilde t_{n}$ with the text embeddings $\tilde t_{n+1}, \tilde t_{n+2}, ...$ to find the next index. This way, we collect the indices of new recipes in our dataset so that we can focus on the revision between these indices.

Moreover, to apply process mining techniques, we built event logs from the writing task. For each group, we collect the activities for each user, by looking at when they submit the first, second and third recipes and all the revision steps in between. 
\subsubsection*{Feature Extraction}
\label{sec:features}

Different aspects of SRL have been researched extensively \cite{flippedclassroms}. In a meta-analysis on online education, \cite{broadbent2015self} found significant associations with academic achievement for five sub-scales of SRL: effort regulation (persistence in learning), time management (ability to plan study time), metacognition (awareness and control of thoughts), critical thinking (ability to carefully examine material), and help-seeking (obtaining assistance if needed)\footnote{The nature of our log data does not allow to represent \textit{help-seeking}.}. Based on these findings, we use the following dimensions to represent student behavior: effort regulation (Number of Revisions, Number of Edits, Time Spent Revising), time management (Time Spent Revising, Pause Times), metacognition (Efficiency, Pause Time), and critical thinking (DIRatio). A detailed description of these feature variables can be found in \cref{sec:detailedFeatures}, \cref{Table:Features}.

\subsubsection*{Building the Learning Path}
Understanding revision behavior implies understanding the underlying process in the writing task (e.g., how long do users in a group take to revise on average or how many users revise). In order to understand this better, process mining, especially process discovery \cite{bogarin}, can help us model and visualize the writing process for users in a group and design a learning path when using adaptive writing support systems \cite{thiemo/processmining}. In this study, we use Directly-Follows Graphs (DFGs) \cite{berti_process_2019}, which represent activities and their relationships\footnote{Other data structures like Petri Nets could also be used. Petri Nets are commonly used to apply process mining \cite{berti_process_2019}.}. This is useful for the field of SRL as it provides a way to visualize and analyze the steps involved in a process, especially revision. A formal definition of DFGs can be found in \cref{sec:dfgdef}.

\section{Results}

\subsubsection*{Revision Strategies}
\begin{figure}[!h]
\Description{bubble plots sorted by the number of revisions}
  \centering
  \includegraphics[width=\linewidth]{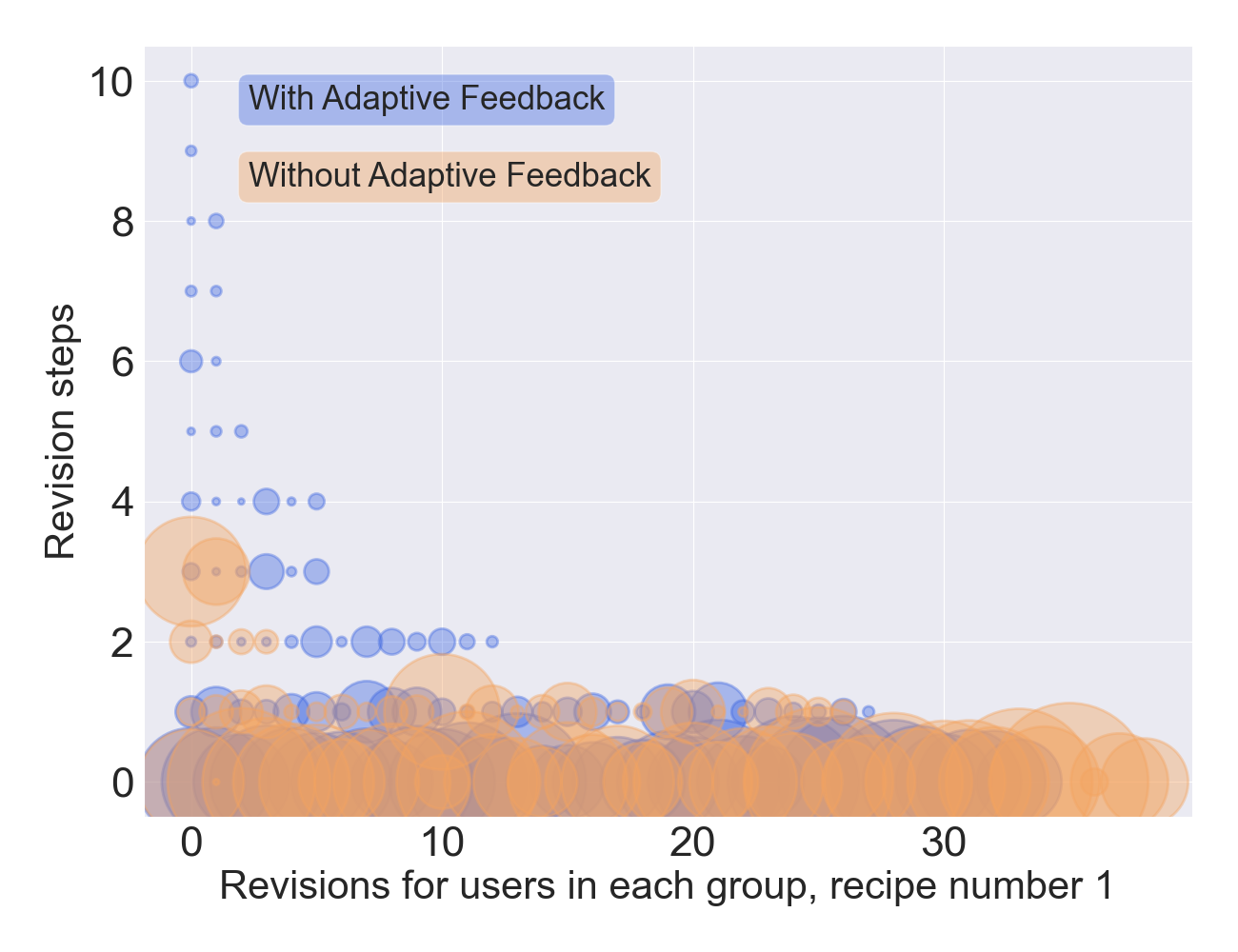}
  \caption{Bubble plot for the first recipe sorted by the number of revisions. The bubbles correspond to the total number of edits (insertions and deletions) for a user at each revision step}
  \label{fig:bubbleplot}
\end{figure}

\begin{figure}[!h]
    \Description{DFGs for the first and second group}
  \centering
  \includegraphics[width=\linewidth]{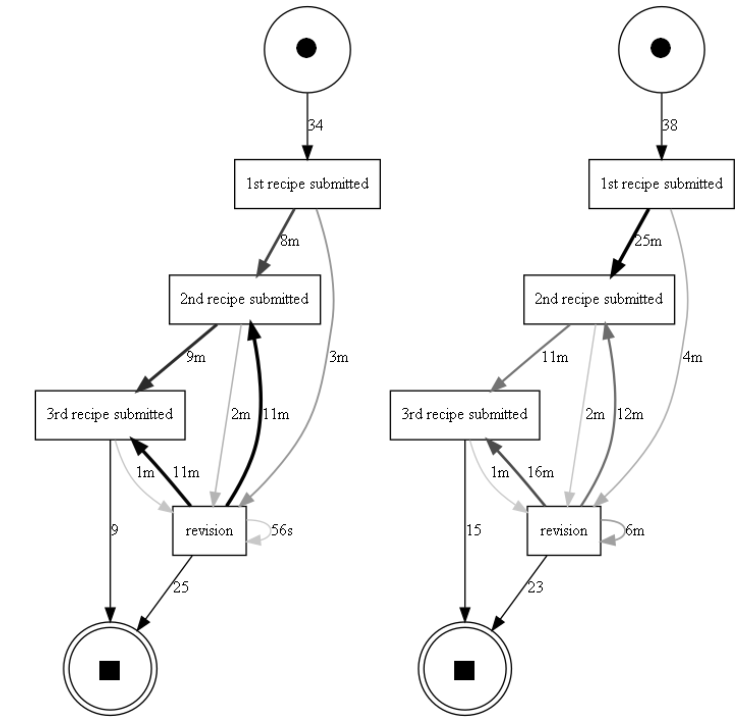}
  \caption{Overview of the SRL behavior of students revising their texts as directly-followed graphs for \texttt{G1} (left) and \texttt{G2} (right) automatically calculated and drawn by our pipeline}
  \label{DFGs}
\end{figure}

With this study, we find that users in different groups revise their texts differently. Recall that \texttt{G1} is given adaptive feedback and \texttt{G2} is not. By providing insightful feedback on what a user can change in their writing, users tend to have more revision steps with fewer edits at each step. However, users not receiving feedback follow the opposite trend, they have fewer revision steps, with a larger number of revisions at each step. This phenomenon is visible in \cref{fig:bubbleplot}. 
In fact, for the first and second texts (\cref{sec:results}, \cref{Table:recipe1,Table:recipe2}), we find $p$-values < 0.05 for Number of Revisions using $t\text{-tests}$, which indicates a significant difference in the number of revisions. This is also underlined by the mean number of revisions and edits. On average, users in \texttt{G1} tend to revise their texts more often, with fewer edits at each step (\cref{sec:results}, \cref{Table:recipe1,Table:recipe2,Table:recipe3}). 
From the directly-follows graphs (\cref{DFGs}), we see that users spend approximately the same amount of time writing recipes and the same amount of time revising at the first revision step. However, we see that users in \texttt{G2} revise much longer when having consecutive revision sessions (6 min on average) compared to \texttt{G1} (56 s)(Fig. \ref{DFGs}). This confirms that users in \texttt{G1} have shorter revision sessions, whereas users in \texttt{G2} have longer revision steps.

\subsubsection*{Engagement}
\begin{figure}[!h]
\Description{user engagement and feature evolution on 4 feature variables over the entirety of the writing experiment}
  \centering
  \hspace{-5pt}\includegraphics[width=1\linewidth]{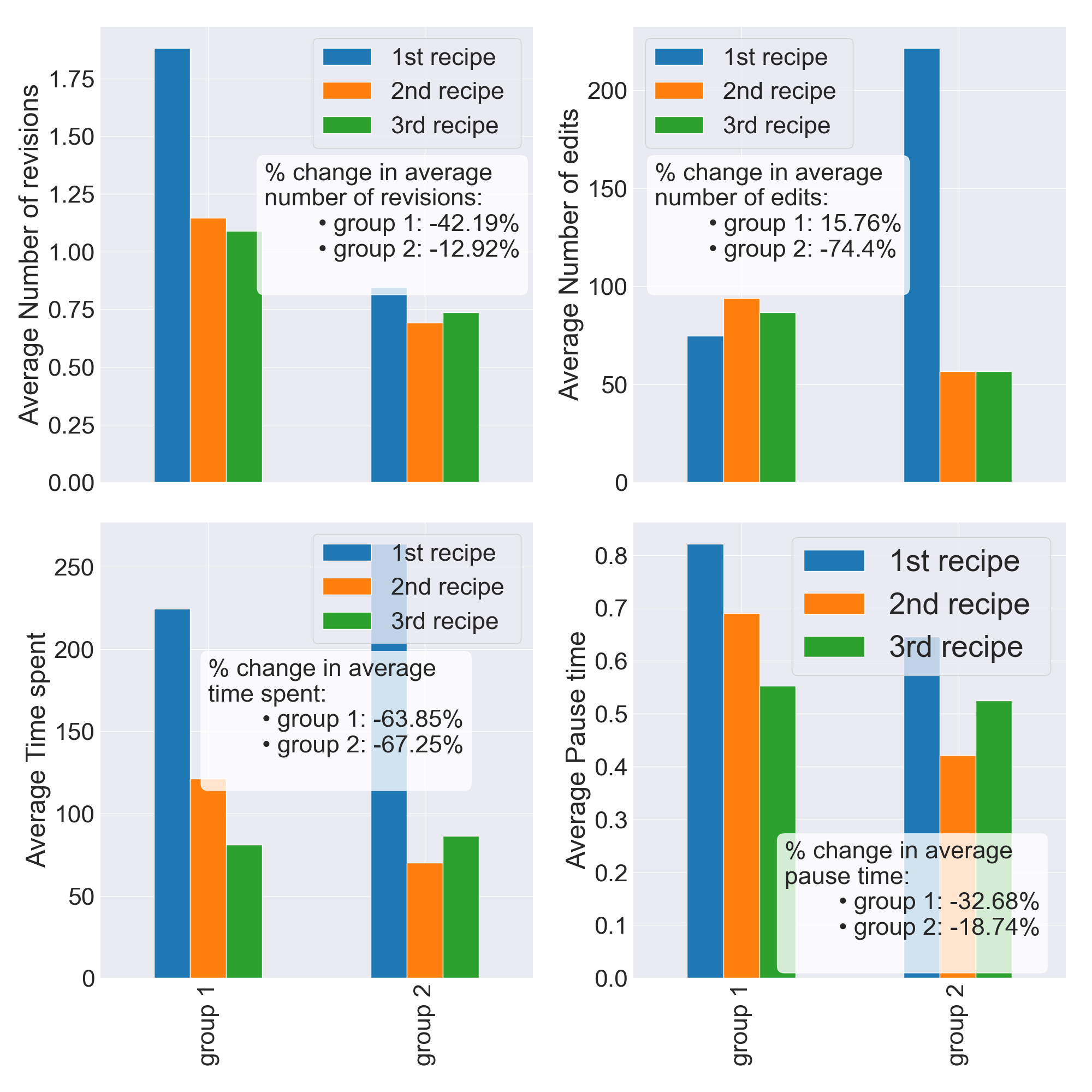}
  \caption{Visualizing user engagement and feature evolution on 4 feature variables over the entirety of the writing experiment}
  \label{fig:summaryplot}
\end{figure}

\begin{figure}[h]
\Description{Percentage of participants that stated that they made no changes when editing their recipes}
    \centering
    \includegraphics[scale=0.35]{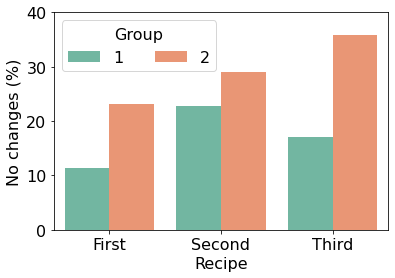}
    \caption{Percentage of participants that stated that they made no changes when editing their recipes in the survey following the experiment}
    \label{fig:changes}
\end{figure}

From the second recipe onwards, we find that users revise less often, perform fewer edits, spend less time revising and type faster (\cref{fig:summaryplot}). This stems from users being less engaged in the task at hand. In fact, users spend 67\% less time revising in \texttt{G2} (\cref{fig:summaryplot}) (from 264 seconds on average for the first recipe to 86.5 seconds for the third one (\cref{Table:recipe1,Table:recipe3})) and 64\% less in \texttt{G1} (from 224 seconds to 81.2). In \texttt{G2}, users perform 74\% fewer edits between the first and last recipe (\cref{fig:summaryplot}). Users in \texttt{G2} performed on average 222 edits when revising for the first recipe and only 57 for the third one (\cref{Table:recipe1,Table:recipe3}). The decrease in pause time for the two groups also declines over time (0.822 to 0.553 seconds on average for \texttt{G1} and 0.646 to 0.525 seconds for \texttt{G2}), even though participants in \texttt{G2} consistently maintain a smaller average pause time when revising.
This is one interpretation of the results and \cref{fig:summaryplot}, another one would be to consider users are improving in this task. On average, pause time for \texttt{G1} decreased by 32.7\% and 18.7\% for \texttt{G2} (\cref{fig:summaryplot}). Shorter pause times indicate better understanding of the task requirements and better task planning \cite{zhang2017investigation}.
This is coherent with the participants' reported changes. As seen in Fig. \ref{fig:changes}, we found that participants from \texttt{G2} increasingly reported making no changes to their recipes (36\% for the third recipe). In contrast, participants in \texttt{G1} continued reporting making changes based on the received adaptive feedback. Nevertheless, there was also an increase in the participants in \texttt{G1} that did not edit the recipe, one participant noted \textit{I didn’t edit as much this time as I remembered to add them the first time around}. 

\subsubsection*{Gender Comparison}
\begin{figure}[!h]
\Description{gender comparison over 2 features}
\centering
    \hspace{-10pt}\includegraphics[scale=0.23]{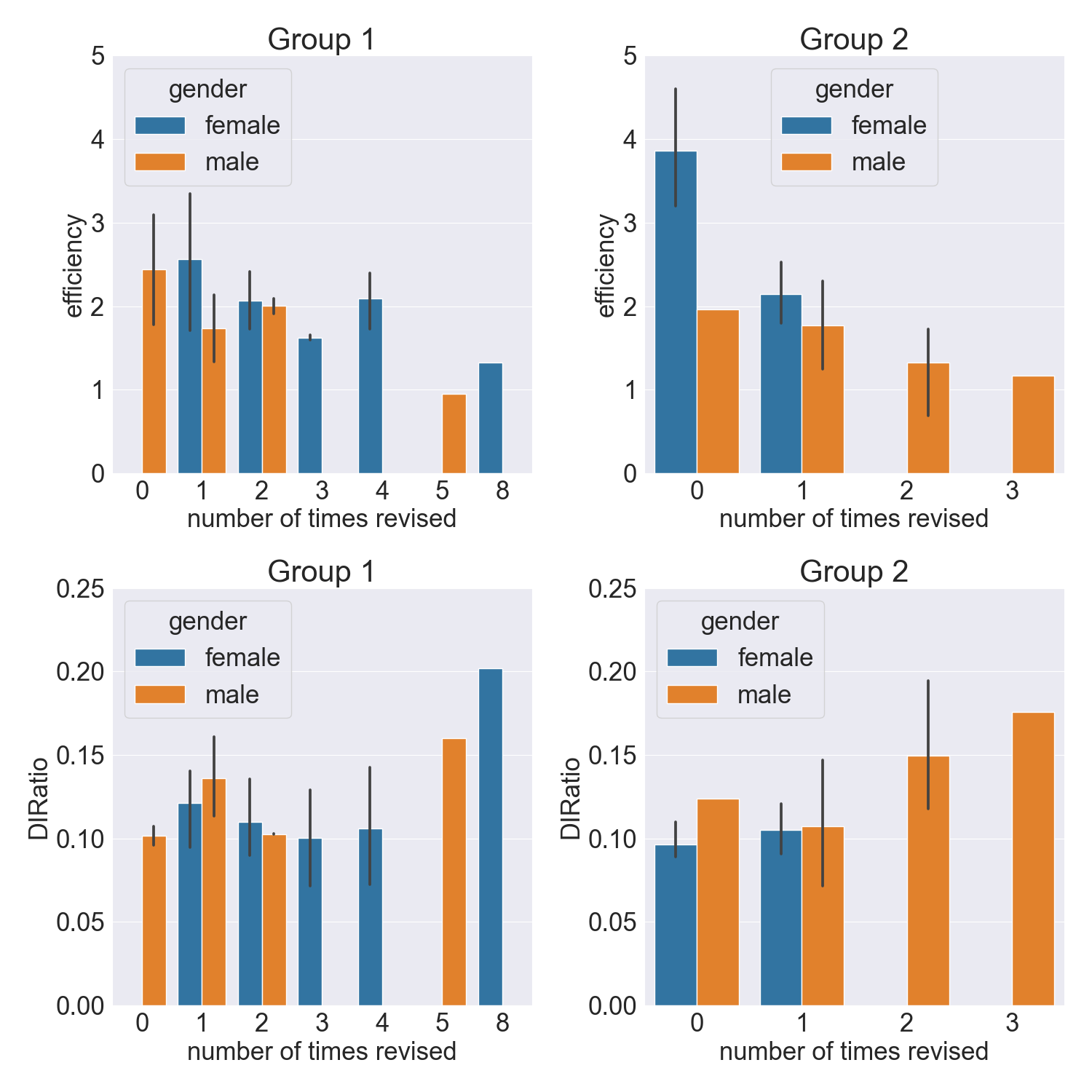}
  \caption{Overview of 2 SRL features from our pipeline comparing males and females}
  \label{fig:gender}
\end{figure}

Research has often found that males tend to be more impaired at composing text in comparison with women. The study in \cite{zhu_analysis_2019} found that female students performed better than male students on a number of levels. Females had higher scores, revised more, and were more efficient: they revised more per unit of time, exhibiting greater writing fluency. In this study, we found that there is a clear distinction in the writing capabilities between males and females. Like \cite{zhu_analysis_2019}, we find females are more efficient in this writing task. They tend to have higher efficiency scores (\cref{fig:gender}) and we find $p=0.0038$ when comparing efficiency scores in \texttt{G2} (\cref{Table:genderP}), which demonstrates the disparity in efficiency distribution between the two groups.
Curiously, males revise less often when receiving feedback (as seen on the x-axis by the number of times revised, (\cref{fig:gender})\footnote{Some outliers were removed (e.g., users who spent over 10'000 seconds revising or users who have very low efficiency scores (few edits over a long period of time)).}). On the contrary, when users do not receive feedback, females revise once at most (because index 0 is not a revision phase, \cref{fig:gender}). This also reinforces women's abilities in their writing, suggesting they feel less need for revision if they do not receive feedback on what they can improve. Regarding the Delete-Insert ratio (DIRatio), although we find there is no statistical difference (\cref{Table:genderP}), we find that males in \texttt{G2} generally have higher scores, especially in \texttt{G2}. Having higher DIRatio scores means users delete a larger portion of their texts (over 15\% for several male users in \texttt{G2}, \cref{fig:gender}). Looking back at SRL, especially on the critical thinking aspect \cite{broadbent2015self}, which is defined as the ability to examine material, we can see males are more self-critical and delete a larger portion of their texts compared to females when they do not receive feedback.

\begin{table}[!h]
\centering
\renewcommand{\arraystretch}{1.3} % Increase the array stretch to add space between rows
\caption{$p$-values for Efficiency and DIRatio features comparing males and females in each group; \textit{*p<0.05, **p<0.01,  ***p<0.001}}
\begin{tabular}{l|>{\centering\arraybackslash}p{2.2cm} >{\centering\arraybackslash}p{2.7cm}}
& With Adaptive Feedback (\texttt{G1}) & Without Adaptive Feedback (\texttt{G2})\\
\hline
Efficiency &  $p=0.215$ & $p=0.0038$** \\
DIRatio &  $p=0.387$ &  $p=0.088$\\
\end{tabular}
\label{Table:genderP}
\end{table}

% and don't forget to end the environment with
% {figure*}, not {figure}!
\section{Discussion \& Conclusion}

With this research, we contribute to the field of understanding the use of intelligent writing systems by learners. We do this by gaining insights into their SRL by inspecting revision behavior.
From the log data we collected, we built and modelled a pipeline to analyze and visualize user behavior in the revision phases of the writing task, by observing different features extracted from the revisions of \texttt{G1} (with adaptive feedback) and \texttt{G2} (without adaptive feedback)(\cref{fig:bubbleplot,fig:summaryplot,fig:gender}). Our analysis revealed that learners in different groups revise using different strategies. Learners who were equipped with adaptive feedback revised more often, with fewer edits at each revision step and users without adaptive feedback followed the opposite trend. 
This suggests that the support provided by the system may influence revision behavior and how it is used. Additionally, we found users seemed to be improving in the writing task as demonstrated by the post-survey and the data, even though they seem to be less engaged from the evolution of the feature variables in \cref{fig:summaryplot}. Finally, we concluded females were more efficient than males in this experiment, by having higher efficiency scores.
While there has been research on the effectiveness of such systems in improving writing skills, there is a limited understanding of how users revise their writing when using these tools. To evaluate users' SRL, it is crucial to have a better understanding of how they self-regulate, especially in writing activities, in order to provide them with the correct tools to improve their writing skills and understand the underlying writing process \cite{bjork2013self, WONG2021106596}.

Regarding future directions, one can focus on clustering revision data in order to gain further insights into the revision behavior in a writing task. We have already done this, by identifying eight reported revisions, including \textit{adding more details}, \textit{changing the structure}, \textit{improving the clarity} or \textit{not making any changes}. Nevertheless, we focus on \textit{not making any changes}, but analyzing other revision reports could help shed light on more differences between the groups. As such, clustering could be used for each group to identify the differences between the two groups or between learners in the same group, to see how users revise when receiving feedback or not.

In conclusion, our research on revision behavior in adaptive writing support systems has shed light on how users in different groups approach revision. The development of a pipeline to study this topic has allowed us to collect and analyze data on user writing and revision activity, leading to the discovery of important patterns and trends. Overall, our study has made a significant contribution to the field by providing a deeper understanding of revision behavior.

%1 pargraph: summary of reserach objective and what we did, also mention dataset and pipline

%3. paragraph repeating the literature gap/ state from 2.1 with citations (3-4 sentences) and then say how we contribute to this 4-5 sentences
%5. pargarph: what we have now on design implications
%
% The following two commands are all you need in the
% initial runs of your .tex file to
% produce the bibliography for the citations in your paper.

\bibliographystyle{abbrv}
\bibliography{sigproc}  % sigproc.bib is the name of the Bibliography in this case

\begin{thebibliography}{10}

\bibitem{allen2016enter}
L.~K. Allen, M.~E. Jacovina, M.~Dascalu, R.~D. Roscoe, K.~M. Kent, A.~D.
  Likens, and D.~S. McNamara.
\newblock Entering the time series space: Uncovering the writing process
  through keystroke analyses.
\newblock {\em International Educational Data Mining Society}, 2016.

\bibitem{dfgs}
A.~Augusto, M.~Dumas, M.~La~Rosa, S.~Leemans, and S.~vanden Broucke.
\newblock Optimization framework for dfg-based automated process discovery
  approaches.
\newblock {\em Software and Systems Modeling}, 20:1--26, 08 2021.

\bibitem{berti_process_2019}
A.~Berti, S.~J. van Zelst, and W.~M.~P. van~der Aalst.
\newblock Process mining for python (pm4py): Bridging the gap between
  process-and data science.
\newblock {\em CoRR}, abs/1905.06169, 2019.

\bibitem{bjork2013self}
R.~A. Bjork, J.~Dunlosky, and N.~Kornell.
\newblock Self-regulated learning: Beliefs, techniques, and illusions.
\newblock {\em Annual review of psychology}, 64:417--444, 2013.

\bibitem{bogarin}
A.~Bogarín, R.~Cerezo, and C.~Romero.
\newblock A survey on educational process mining.
\newblock {\em Wiley Interdisciplinary Reviews: Data Mining and Knowledge
  Discovery}, 8, 09 2017.

\bibitem{broadbent2015self}
J.~Broadbent and W.~L. Poon.
\newblock Self-regulated learning strategies \& academic achievement in online
  higher education learning environments: A systematic review.
\newblock {\em The Internet and Higher Education}, 27:1--13, 2015.

\bibitem{CHEN201859}
X.~Chen, L.~Breslow, and J.~DeBoer.
\newblock Analyzing productive learning behaviors for students using immediate
  corrective feedback in a blended learning environment.
\newblock {\em Computers \& Education}, 117:59--74, 2018.

\bibitem{DBLP:coenen}
A.~Coenen, L.~Davis, D.~Ippolito, E.~Reif, and A.~Yuan.
\newblock Wordcraft: a human-ai collaborative editor for story writing.
\newblock {\em CoRR}, abs/2107.07430, 2021.

\bibitem{Bandura}
J.~de~la Fuente, J.~M. Martínez-Vicente, F.~H. Santos, P.~Sander, S.~Fadda,
  E.~Karagiannopoulou, E.~Boruchovitch, and D.~F. Kauffman.
\newblock Advances on self-regulation models: A new research agenda through the
  sr vs er behavior theory in different psychology contexts.
\newblock {\em Frontiers in Psychology}, 13, 2022.

\bibitem{emig1971composing}
J.~Emig.
\newblock The composing processes of twelfth graders.
\newblock 1971.

\bibitem{10.2307/356600}
L.~Flower and J.~R. Hayes.
\newblock A cognitive process theory of writing.
\newblock {\em College Composition and Communication}, 32(4):365--387, 1981.

\bibitem{gero-etal-2022-design}
K.~Gero, A.~Calderwood, C.~Li, and L.~Chilton.
\newblock A design space for writing support tools using a cognitive process
  model of writing.
\newblock In {\em Proceedings of the First Workshop on Intelligent and
  Interactive Writing Assistants (In2Writing 2022)}, pages 11--24, Dublin,
  Ireland, May 2022. Association for Computational Linguistics.

\bibitem{grootendorst2022bertopic}
M.~Grootendorst.
\newblock Bertopic: Neural topic modeling with a class-based tf-idf procedure.
\newblock {\em arXiv preprint arXiv:2203.05794}, 2022.

\bibitem{hayes2012modeling}
J.~R. Hayes.
\newblock Modeling and remodeling writing.
\newblock {\em Written communication}, 29(3):369--388, 2012.

\bibitem{2022arXiv220106796L}
M.~{Lee}, P.~{Liang}, and Q.~{Yang}.
\newblock {CoAuthor: Designing a Human-AI Collaborative Writing Dataset for
  Exploring Language Model Capabilities}.
\newblock {\em arXiv e-prints}, page arXiv:2201.06796, Jan. 2022.

\bibitem{leijten_keystroke_2013}
M.~Leijten and L.~Van~Waes.
\newblock Keystroke {Logging} in {Writing} {Research}: {Using} {Inputlog} to
  {Analyze} and {Visualize} {Writing} {Processes}.
\newblock {\em Written Communication}, 30(3):358--392, July 2013.
\newblock Publisher: SAGE Publications Inc.

\bibitem{mccutchen1996capacity}
D.~McCutchen.
\newblock A capacity theory of writing: Working memory in composition.
\newblock {\em Educational psychology review}, 8(3):299--325, 1996.

\bibitem{flippedclassroms}
P.~Mejia-Domenzain, M.~Marras, C.~Giang, and T.~K\"{a}ser.
\newblock Identifying and comparing multi-dimensional student profiles across
  flipped classrooms.
\newblock In {\em Artificial Intelligence in Education: 23rd International
  Conference, AIED 2022, Durham, UK, July 27–31, 2022, Proceedings, Part I},
  page 90–102, Berlin, Heidelberg, 2022. Springer-Verlag.

\bibitem{nova2018utilizing}
M.~Nova.
\newblock Utilizing grammarly in evaluating academic writing: A narrative
  research on efl students’ experience.
\newblock {\em Premise: Journal of English Education and Applied Linguistics},
  7(1):80--96, 2018.

\bibitem{DBLP:padmakumar}
V.~Padmakumar and H.~He.
\newblock Machine-in-the-loop rewriting for creative image captioning.
\newblock {\em CoRR}, abs/2111.04193:573--586, 2021.

\bibitem{pennington2014glove}
J.~Pennington, R.~Socher, and C.~D. Manning.
\newblock Glove: Global vectors for word representation.
\newblock In {\em Proceedings of the 2014 conference on empirical methods in
  natural language processing (EMNLP)}, pages 1532--1543, 2014.

\bibitem{DBLP:conf/emnlp/ReimersG19}
N.~Reimers and I.~Gurevych.
\newblock Sentence-bert: Sentence embeddings using siamese bert-networks.
\newblock In K.~Inui, J.~Jiang, V.~Ng, and X.~Wan, editors, {\em Proceedings of
  the 2019 Conference on Empirical Methods in Natural Language Processing and
  the 9th International Joint Conference on Natural Language Processing,
  {EMNLP-IJCNLP} 2019, Hong Kong, China, November 3-7, 2019}, pages 3980--3990.
  Association for Computational Linguistics, 2019.

\bibitem{sinharay_prediction_2019}
S.~Sinharay, M.~Zhang, and P.~Deane.
\newblock Prediction of {Essay} {Scores} {From} {Writing} {Process} and
  {Product} {Features} {Using} {Data} {Mining} {Methods}.
\newblock {\em Applied Measurement in Education}, 32(2):116--137, Apr. 2019.
\newblock Publisher: Routledge \_eprint:
  https://doi.org/10.1080/08957347.2019.1577245.

\bibitem{stallard1974analysis}
C.~K. Stallard.
\newblock An analysis of the writing behavior of good student writers.
\newblock {\em Research in the Teaching of English}, 8(2):206--218, 1974.

\bibitem{vandermeulen2020reporting}
N.~Vandermeulen, M.~Leijten, and L.~Van~Waes.
\newblock Reporting writing process feedback in the classroom using keystroke
  logging data to reflect on writing processes.
\newblock {\em Journal of Writing Research}, 12(1):109--139, 2020.

\bibitem{WAMBSGANSS2022104644}
T.~Wambsganss, A.~Janson, and J.~M. Leimeister.
\newblock Enhancing argumentative writing with automated feedback and social
  comparison nudging.
\newblock {\em Computers \& Education}, 191:104644, 2022.

\bibitem{wambsganss2022alen}
T.~Wambsganss, T.~Kueng, M.~S{\"o}llner, and J.~M. Leimeister.
\newblock Arguetutor: An adaptive dialog-based learning system for
  argumentation skills.
\newblock {\em Proceedings of the 2021 CHI Conference on Human Factors in
  Computing Systems}, 2021.

\bibitem{wambsganss__2020}
T.~Wambsganss, C.~Niklaus, M.~Cetto, M.~Söllner, S.~Handschuh, and J.~M.
  Leimeister.
\newblock {AL}: an adaptive learning support system for argumentation skills.
\newblock In {\em Proceedings of the 2020 {CHI} Conference on Human Factors in
  Computing Systems}, pages 1--14.

\bibitem{wambsganss2021supporting}
T.~Wambsganss, C.~Niklaus, M.~S{\"o}llner, S.~Handschuh, and J.~M. Leimeister.
\newblock Supporting cognitive and emotional empathic writing of students.
\newblock In {\em The Joint Conference of the 59th Annual Meeting of the
  Association for Computational Linguistics and the 11th International Joint
  Conference on Natural Language Processing (ACL-IJCNLP 2021)}, volume
  abs/2105.14815, 2021.

\bibitem{thiemo/processmining}
T.~Wambsganß, A.~Schmitt, T.~Mahnig, A.~Ott, N.~Ngo, J.~Geyer-Klingeberg,
  J.~Nakladal, and J.~M. Leimeister.
\newblock The potential of technology-mediated learning processes: A taxonomy
  and research agenda for educational process mining.
\newblock 10 2021.

\bibitem{WONG2021106596}
J.~Wong, M.~Baars, B.~B. {de Koning}, and F.~Paas.
\newblock Examining the use of prompts to facilitate self-regulated learning in
  massive open online courses.
\newblock {\em Computers in Human Behavior}, 115:106596, 2021.

\bibitem{UCAM-CL-TR-904}
Z.~Yuan.
\newblock {Grammatical error correction in non-native English}.
\newblock Technical Report UCAM-CL-TR-904, University of Cambridge, Computer
  Laboratory, Mar. 2017.

\bibitem{zhang2016classification}
M.~Zhang, J.~Hao, C.~Li, and P.~Deane.
\newblock Classification of writing patterns using keystroke logs.
\newblock In L.~A. van~der Ark, D.~M. Bolt, W.-C. Wang, J.~A. Douglas, and
  M.~Wiberg, editors, {\em Quantitative Psychology Research}, pages 299--314,
  Cham, 2016. Springer International Publishing.

\bibitem{zhang2017investigation}
M.~Zhang, D.~Zou, A.~Wu, P.~Deane, C.~Li, B.~Zumbo, and A.~Hubley.
\newblock An investigation of the writing processes in timed task condition
  using keystrokes.
\newblock {\em Understanding and investigating response processes in validation
  research}, pages 321--339, 2017.

\bibitem{zhu_analysis_2019}
M.~Zhu, M.~Zhang, and P.~Deane.
\newblock Analysis of {Keystroke} {Sequences} in {Writing} {Logs}.
\newblock {\em ETS Research Report Series}, 2019(1):1--16, 2019.
\newblock \_eprint: https://onlinelibrary.wiley.com/doi/pdf/10.1002/ets2.12247.

\bibitem{zimmerman}
B.~Zimmerman.
\newblock Becoming a self-regulated learner: An overview.
\newblock {\em Theory Into Practice}, 41:64--70, 06 2002.

\end{thebibliography}
% You must have a proper ".bib" file
%  and remember to run:
% latex bibtex latex latex
% to resolve all references
%
%APPENDICES are optional
%\balancecolumns

\appendix
%Appendix A

\section{Features}
\label{sec:detailedFeatures}
\cref{Table:Features} gives a detailed description of the features we study in this paper, extracted in our pipeline. 
\begin{table*}
\centering
\caption{Overview of feature variables automatically calculated through our pipeline to measure SRL behavior of users in their writing exercises based on keystroke logs}
\renewcommand{\arraystretch}{2} % Increase the array stretch to add space between rows

\begin{tabular}{p{0.2\linewidth} p{0.75\linewidth}}
\textbf{Feature Variables} & \textbf{Description} \\
\hline
Number Of Revisions & For each user, we count the amount of times they revise each time they write a recipe (i.e., when they submitted then re-edited their texts). This gives a sense of the effort put into the revision phase of the writing task. \\
\hline
Number of edits & The total number of insertions and deletions during a revision step. Insertions are counted as any characters that are typed including whitespaces, and deletions are counted as the number of times the user presses any of the Backspace or Delete buttons. \\
\hline
Time Spent Revising in seconds & We compute the average time users spend revising for each group, for each recipe. This allows us to compare the two groups and to estimate the effort put in by both groups.\\
\hline
Delete-Insert Ratio (DIRatio) & The average deletions over insertions ratio, which approximately captures the extent of editing and revision of any kind \cite{zhu_analysis_2019}.\\
\hline
Efficiency & Estimated by the number of insertions per second, which indicates a general writing speed. This feature is arguably an indicator of writing fluency \cite{zhu_analysis_2019}. \\
\hline
Pause Time during Revision in seconds & For each user, we collect the inter-key time interval and compute the mean of these intervals. This captures the average lag time between two adjacent keystroke actions \cite{zhu_analysis_2019}. This feature captures the effort and persistence level of users. \\
\hline
\end{tabular}

\label{Table:Features}
\end{table*}

\section{Directly-Follows Graphs}
\label{sec:dfgdef}
In process mining, a Directly-Follows Graph is a directed graph that represents the sequence of activities in a process based on event logs. Formally, given a set of activities $\mathcal{A}$, an event log $\mathcal{L}$ is a multiset of traces $t\in \mathcal{L}$, where $t$ is a sequence of activities $t=(a_1, a_2,\;..., a_n)$, with $a_i\in \mathcal{A}$  $1\leq i \leq n$  \cite{dfgs}.
Given the event log $\mathcal L$, the DFG $\mathcal G$ is a directed graph such that $\mathcal{G} = (V,E)$. $V$ is the set of activities in $\mathcal{L}$: $V = \{a\in \mathcal{A}\;|\; \exists t\in \mathcal{L} \land a \in t \}$. $E$ is defined as $E=\{(u,v) \in V\times{V}\;|\; \exists t =(a_1, a_2, ..., a_n), \; t\in \mathcal{L} \; \land \; a_i = u \; \land \; a_{i+1}\}$ \cite{dfgs}.

\section{Separating writing sessions}
\label{sec:writingsessions}
\cref{alg:indices} describes our implementation of session separation. Each participant wrote a first version of their recipes, then revised it, before starting the next recipes. To focus on the revision sessions, we needed to implement a function which captures the indices in the dataset where participants started their recipes. First, we preprocess the submitted text by removing noisy characters, such as punctuation and return the list of sanitized words. Then we use the GloVe model to convert the words to vectors and return one 50-dimensional vector which is the sum of each word embedding. Then we recursively find the indices of new recipes using cosine distance. However, the algorithm is 91\% accurate: this is because sometimes users submitted random strings or the revisions led to the algorithm detecting another recipe. We adjusted the missing indices by hand by looking at the dataset. 

\begin{algorithm}
\caption{Separating writing sessions using cosine distance}

\begin{algorithmic}
\Function{separateSessions}{}
\State $model \gets \textsc{GloVeModel}$   
\Function{getVector}{$text$}
\State $p \gets \textsc{preProcess}(text)$ \Comment{splits and sanitizes $text$}
\State $arr\gets$ Initialize an empty list 
\For{$word \in p$}
\State add $model[word]$ to $arr$ \textbf{if} $word \in model$ 

\EndFor
\State \Return $\textsc{np.sum}(arr, axis=0)$ \Comment{uses numpy}
\EndFunction
\\

\Function{computeIndices}{$startIndex, accumulator$}
    \State $recipes \gets$ retrieve $recipes$ from the dataset
    \State $size \gets $ the total number of $recipes$
    \If{$startIndex \geq size - 1$}
        \State \Return $accumulator$
    \EndIf
    \State $vec \gets \textsc{getVector}(recipes[startIndex])$
    \For{$n \gets startIndex$ to $size$}
        \State $d \gets 1 - \textsc{cosineDist}(vec, \textsc{getVector}(recipes[n]))$
        \If{$d < 0.995$}
            \State add $n$ to the $accumulator$
            \State \Return \textsc{computeIndices}($n$, $accumulator$)
        \EndIf
    \EndFor
\EndFunction
\State \Return \textsc{computeIndices}($0$, empty $accumulator$)

\EndFunction
\end{algorithmic}
\label{alg:indices}
\end{algorithm}

\section{Results}
\label{sec:results}
\subsubsection*{Detailed Results}
\cref{Table:recipe1,Table:recipe2,Table:recipe3} report the mean, standard deviation of different feature variables for both groups, as well as $p$-values.

\begin{table*}[!h]
\centering
 \caption{Overview of SRL features from our pipeline for the first written text 
 between students receiving adaptive feedback (\texttt{G1}) and no feedback (\texttt{G2}) based on our data set}
\begin{tabular}{l|p{2cm}|c|p{2.2cm}|c|c}
 &\multicolumn{2}{c}{With Adaptive Feedback (\texttt{G1})} & \multicolumn{2}{|c|}{Without Adaptive Feedback (\texttt{G2})} &\\
 \hline
  Feature Variables & \centering Mean & Std & \centering Mean & Std & $p$-values\\
 \hline
 Number of Revisions &\centering 1.882 & 2.166 & \centering0.846 & 0.735 & 0.0071\\
 Number of Edits & \centering75.734 & 95.114 & \centering222.73 & 338.574 & 0.24 \\
 Time Spent Revising (sec) & \centering224.48 & 237.37 & \centering264.01 & 530.47 & 0.694\\
 Pause Time in Revision (sec) & \centering 0.822 & 0.225 & \centering 0.646 & 0.08 & 0.339
\end{tabular}
 
  \label{Table:recipe1}
\end{table*}

\begin{table*}[!h]
\centering
\caption{Overview of SRL features from our pipeline for the second written text 
 between students receiving adaptive feedback (\texttt{G1}) and no feedback (\texttt{G2}) based on our data set}
  
\begin{tabular}{l|p{2cm}|c|p{2.2cm}|c|c}
 &\multicolumn{2}{c}{With Adaptive Feedback (\texttt{G1})} & \multicolumn{2}{|c|}{Without Adaptive Feedback (\texttt{G2})} &\\
 \hline
 Feature Variables & \centering{Mean} & Std & \centering{Mean} & Std & $p$-values\\
 \hline
 Number of Revisions & \centering1.147 & 1.033 & \centering 0.692 & 0.722 & 0.033\\
 Number of Edits & \centering95.051 & 192.92 & \centering 57.52 & 61.31 & 0.26 \\
 Time Spent Revising (sec) & \centering121.19 & 148 & \centering 70.15 & 120.9 & 0.11\\
 Pause Time in Revision (sec) & \centering 0.69 & 0.365 & \centering 0.421 & 0.2 & 0.089
\end{tabular}
  \label{Table:recipe2}
\end{table*}

\begin{table*}[!h]
\centering
 \caption{Overview of SRL features from our pipeline for the third written text 
 between students receiving adaptive feedback (\texttt{G1}) and no feedback (\texttt{G2}) based on our data set}
  
\begin{tabular}{l|p{2cm}|c|p{2.2cm}|c|c}
 &\multicolumn{2}{c}{With Adaptive Feedback (\texttt{G1})} & \multicolumn{2}{|c|}{Without Adaptive Feedback (\texttt{G2})} &\\
 \hline
 Feature Variables & \centering Mean & Std & \centering Mean & Std & $p$-values\\
 \hline
 Number of Revisions & \centering1.12 & 1.05 & \centering0.737 & 0.676 & 0.11\\
 Number of Edits & \centering87.61 & 270.75 & \centering57.7
  & 71.7 & 0.45\\
 Time Spent Revising (sec) & \centering81.2 & 115.8 & \centering86.5 & 231.9 & 0.905\\
 Pause Time in Revision (sec) &\centering 0.553 & 0.212 & \centering 0.525 & 0.296 & 0.85
\end{tabular}
 \label{Table:recipe3}
\end{table*}

% This next section command marks the start of
% Appendix B, and does not continue the present hierarchy

\balancecolumns
% That's all folks!
\end{document}